 \documentclass[final,1p,times]{elsarticle}

\usepackage{amssymb}
\usepackage{subfig}
\usepackage{amsthm, amsmath}
\usepackage{floatrow}
\usepackage{wrapfig, multirow}
\usepackage{hhline}

\newfloatcommand{capbtabbox}{table}[][\FBwidth]

\begin{document}

\begin{frontmatter}






\title{Med7: a transferable clinical natural language processing model for electronic health records}


\author[1]{Andrey Kormilitzin\fnref{myfootnote}}
\author[1]{Nemanja Vaci}
\author[1]{Qiang Liu}
\author[1]{Alejo Nevado-Holgado}
\address[1]{Department of Psychiatry, Warneford Hospital, Oxford, OX3 7JX, UK}
\fntext[myfootnote]{Corresponding author: andrey.kormilitzin@psych.ox.ac.uk}

\begin{abstract}

The field of clinical natural language processing has been advanced significantly since the introduction of deep learning models. The self-supervised representation learning and the transfer learning paradigm became the methods of choice in many natural language processing application, in particular in the settings with the dearth of high quality manually annotated data. Electronic health record systems are ubiquitous and the majority of patients' data are now being collected electronically and in particular in the form of free text. Identification of medical concepts and information extraction is a challenging task, yet important ingredient for parsing unstructured data into structured and tabulated format for downstream analytical tasks. In this work we introduced a named-entity recognition model for clinical natural language processing. The model is trained to recognise seven categories: drug names, route, frequency, dosage, strength, form, duration. The model was first self-supervisedly pre-trained by predicting the next word, using a collection of 2 million free-text patients' records from MIMIC-III corpora and then fine-tuned on the named-entity recognition task. The model achieved a lenient (strict) micro-averaged F1 score of 0.957 (0.893) across all seven categories. Additionally, we evaluated the transferability of the developed model using the data from the Intensive Care Unit in the US to secondary care mental health records (CRIS) in the UK. A direct application of the trained NER model to CRIS data resulted in reduced performance of F1=0.762, however after fine-tuning on a small sample from CRIS, the model achieved a reasonable performance of F1=0.944. This demonstrated that despite a close similarity between the data sets and the NER tasks, it is essential to fine-tune on the target domain data in order to achieve more accurate results. 

\end{abstract}




\end{frontmatter}


\section{Introduction}
\label{intro}

Recent years have seen remarkable technological advances in digital platforms for medicine and healthcare. The majority of patients' medical records are now being collected electronically and represent unparalleled opportunities for research, delivering better health care and improving patients' outcomes. However, a substantial amount of patients' information is contained in a free-text form as summarised by clinicians, nurses and caregivers through the interview and assessments. The textual medical records contain rich information about a patient's history as it is expressed in natural language and allows to reflect nuanced details. However, free-texts pose certain challenges in their direct utilisation as opposed to structured and ready-to-use data sources. Manual processing of all patients' free-texts records severely limits the utilisation of unstructured data and makes the process of data mining extremely expensive. On the other hand, machine learning algorithms are well poised to process a large amount of data, spot unusual interactions and extract meaningful information. Recent lines of research in the field of natural language processing (NLP), such as deep contextualised word representations \cite{peters2018deep}, Transformer-based architectures \cite{vaswani2017attention} and large language models \cite{devlin2018bert}, offer new opportunities for clinical natural language processing with unstructured medical records \cite{velupillai2018using}. However, despite recent technological advances, there are a number of challenges pertinent to the field of clinical NLP which should be addressed in order to develop trustworthy models for information extraction. One of the foremost challenges is the dearth of high quality annotated examples to robustly train generalisable models. Large amounts of medical data cannot be made publicly available for crowdsourcing annotations, similar to ImageNet \cite{imagenet_cvpr09, DBLP:journals/corr/LinMBHPRDZ14} or by means of Amazon Mechanical Turk, due to ethical consideration of patients’ privacy preservation and information security \cite{entzeridou2018public}. Therefore, most of the data annotations are made by a limited amount of domain experts, such as clinicians or nurses, and cannot be shared. Since 2006, the Informatics for Integrating Biology and the Bedside (i2b2) initiative \cite{uzuner2008identifying} has been organising regular challenges on clinical natural language processing and the organisers have been providing a sample of carefully selected for each particular task fully anonymised annotated data. The data were sourced from the Medical Information Mart for Intensive Care (MIMIC) electronic health records (EHR) database \cite{johnson2016mimic}. 

Identification of concepts of interest in free-texts is a sub-task of information extraction (IE), more commonly known as named-entity recognition (NER). The NER task seeks to classify words into predefined categories \cite{schutze2008introduction} and to assign labels to them. A robust and accurate NER model for identification of medical concepts, such as drug names, strength, frequency of administration, reported symptoms, diagnoses, health score and many more, is an essential and foundational component of any clinical IE system. However, in order to develop a reliable and generalisable NER model for real-world observational data, one should first address a number of challenges.

Despite the availability of both, annotated i2b2 data and the entire MIMIC electronic health records (EHR) database, the models developed using these data sources are not guaranteed to generalise robustly on other, yet similar EHR datasets, even on the same downstream tasks. Many supervised learning algorithms are based on the assumption that the training and test sets are sampled from the same distribution. However, when the target and the source domains are different, it is expected that the model will underperform \cite{patel2015visual}. One of the potential solutions to transferability of a model trained to recognise concepts from labelled data in a source domain that also performs well on a different but related target domain, regarded as domain adaptation \cite{ganin2016domain, goodfellow2016deep}.

In this study we address the aforementioned problems by implementing three strategies.  First, the underlying deep neural network language model was self-supervisedly pre-trained on the entire MIMIC-III corpora comprising more than 2 million documents using the cloze-style approach \cite{baevski2019cloze}. Second, using the weak-supervision method \cite{ratner2019role}, we developed synthetic training data with noisy labels. Lastly, we synergistically incorporated all ingredients into an active learning with human-in-the-loop approach to maximise the accuracy of the NER model. 

Additionally, we demonstrated that the developed NER model trained on a source domain from the intensive care MIMIC EHR data in the US, failed to generalise well on the target domain sourced from the secondary care mental health Clinical Record Interactive Search system (CRIS) in the United Kingdom. We also showed that using domain adaption, the NER model could be adequately transferred from MIMIC to CRIS.

\section{Related work}
\label{rel_work}

The topic of clinical natural language processing and information extraction has been actively researched over the past years, in particular with the introduction and adoption of electronic health records platforms. The methods have evolved from simple logic and rule-based systems to complex deep learning architectures \cite{dalianis2018clinical, wu2020deep}. One of the common approaches to information extraction is by transforming free text data into coded representation via lookup tables, such as universal medical language system (UMLS) \cite{bodenreider2004unified} or structured clinical vocabulary for use in an electronic health record (SNOMED CT). Some rule-based systems used semantic lexicons to identify concepts in biomedical literature \cite{gurulingappa2012development} with more complex linguistic features. With the advances in machine learning algorithms, such methods as hidden Markov models and conditional random fields \cite{zhou2002named} were used to label entities for the NER task. In in last decade, deep learning methods have played an essential role in developing more capable models for natural language processing and in particular, in the biomedical domain. Word embeddings \cite{mikolov2013distributed, pennington2014glove} were introduced as numerical representation of textual data and were used as input layers to deep neural networks. For a comprehensive review on word embeddings for clinical applications please refer to \cite{ks2019secnlp}. More recently, the unsupervised model pre-training on a large collection of unlabelled data with further fine-tuning on a downstream task, has taken off and demonstrated its high potential \cite{howard2018universal}. Since the introduction of the Transformer-based deep neural network architectures, such as BERT \cite{devlin2018bert}, Roberta \cite{liu2019roberta}, XLNet \cite{yang2019xlnet} and others, the transfer learning approach of reusing pre-trained models became the method of choice for the majority of NLP tasks. Some notable examples of pre-trained deep learning models for biomedical natural language processing are: BioBERT \cite{lee2020biobert} for text-mining, ClinicalBERT \cite{huang2019clinicalbert, alsentzer2019publicly} for contextual word representations fine-tuned on the electronic health records and predicting hospital readmission. Another open source Python library 'scispaCy' \cite{neumann2019scispacy} was recently introduced for biomedical natural language processing. In this work we developed an open source named-entity recognition model dedicated to identification of seven categories related to medications mentioned in free-text electronic patient records.

\section{Materials and Methods}
\label{data_methods}

\subsection{Data}

The annotated data set was sourced from MIMIC-III (Medical Information Mart for Intensive Care-III) electronic health records data base \cite{johnson2016mimic} as part of the Track 2 of The 2018 National NLP Clinical Challenges (n2c2) Shared Task on drug related concepts extraction, including adverse drug events (ADE) and reasons for prescription \cite{henry20192018}. The data set comprised a collection of discharge letters from the Intensive Care Unit (ICU) and contained very rich and detailed information about medications used for treatment. The data set was randomly split and provided by the organisers into training and test sets with 303 and 202 documents respectively. The documents were annotated for nine categories: ADE, Dosage, Drug, Duration, Form, Frequency, Reason, Route and Strength. For the purpose of the current work we considered only seven drug-related categories and discarded two categories such as ADE and Reason. We aimed to develop a model for medications and their related information extraction which will be beneficial to biomedical community and be robustly used in a variety of downstream nature language processing tasks using free text medical records. The description of the data sets and annotation statistics are summarised in Table \ref{tab:annotated_summary_stats}.

\begin{table}[htb]
    \centering
    \begin{tabular}{l|r|r|r}
        Types of annotated entities   &   Train   &   Test    &   Total   \\ \hline\hline
        Dosage          &   4227    &   2681    &   6908    \\
        Drug            &   16257   &   10575   &   26832   \\
        Duration        &   592     &   378     &   970     \\        
        Form            &   6657    &   4359    &   11016   \\
        Frequency       &   6281    &   4012    &   10293   \\
        Route           &   5460    &   3513    &   8973    \\
        Strength        &   6694    &   4230    &   10924   \\\hline        
        Number of documents & 303   &   202     &   505     \\
        Total number of words &   957972  &  627771  &  1585743         \\
        Total number of unique words &   27602  &    21729       &    35763       \\ \hline
    \end{tabular}
    \caption{Distribution of gold-annotated entities and text summary statistics of the training and test data sets. The number of unique tokens is computed by lowercasing words.}
    \label{tab:annotated_summary_stats}
\end{table}

In addition to MIMIC-III and 2018 n2c2 data sets, we evaluated the developed model on electronic medical records sourced from the Clinical Record Interactive Search (UK-CRIS) platform, which is the largest secondary care mental health database in the United Kingdom. UK-CRIS contains more than 500 million clinical notes from 2.7 million de-identified patient records from 12 National Health Service (NHS) Network Partners across the UK \footnote{https://crisnetwork.co}.

\subsection{Methods}

\subsubsection{Text pre-processing}

In order to compare the performance of the developed medication extraction model using MIMIC-III (n2c2 2018) and UK-CRIS data, basic text cleaning and pre-processing steps were taken to standardise texts. UK-CRIS notes that were uploaded as scanned documents and transformed into electronic texts via optical character recognition (OCR) process, were cleaned from such artefacts as email addresses, non-ASCII characters, website URLs, HTML or XML tags. Additionally, standard escape sequences ('\textbackslash t', '\textbackslash n' and '\textbackslash r') were also removed and the offsets of gold-annotated entities were adjusted accordingly.

\subsubsection{Self-supervised learning}
\label{sec:self-supervision}
The main obstacle to developing an accurate information extraction model is the dearth of a sufficient amount of high-quality annotated data to train the model. In contrast to publicly available large manually annotated data sets for computer vision \cite{imagenet_cvpr09, DBLP:journals/corr/LinMBHPRDZ14} and for various natural language processing downstream tasks \cite{10.5555/1614049.1614064, 2016arXiv160605250R, maas-EtAl:2011:ACL-HLT2011} manually annotated texts for clinical concepts extraction are quite rare \cite{henry20192018}. The shortage of annotated clinical data is mainly due to privacy concerns and potential identification of personal medical information of patients. Several lines of research have addressed the problem of learning from limited annotated data in the clinical domain \cite{hofer2018few, gligic2020named, wang2019clinical} and pre-training of the underlying language model and word representations generally leads to better performance with less data \cite{howard2018universal}. 

In this work, we used the spaCy's \footnote{https://spacy.io} implementation of a cloze-style word reconstruction, similar to the masked language model objectives introduced in BERT \cite{devlin2018bert}, but instead of predicting the exact word identifier from the vocabulary, the GloVe \cite{pennington2014glove} word's vector was predicted using a static embedding table with a cosine loss function. The pre-trained language model was then used to initialise the weights of convolutional neural network layers, rather than starting with random weights. We experimented with various combinations of hyperparameters of the language model, such as the number of rows and width of embedding tables and a depth of convolutional layers.

\subsubsection{Named entity recognition model}

The task of locating concepts of interest in unstructured text and their subsequent classification into predefined categories, for example: drug names, dosages or frequency of administration is a sub-task of information extraction and called named-entity recognition (NER). There are various implementations of NER systems, ranging from rule-based string matching approaches \cite{schutze2008introduction} to complex Transformer models \cite{vaswani2017attention} or their hybrid combinations. In this work the named-entity recognition model for extraction of medication information was implemented in Python 3.7 using spaCy open source library for NLP tasks \cite{spacy2}. Although there exists a good number of NLP libraries, such as: NLTK \cite{BirdKleinLoper09}, NLP4J \cite{choi2016dynamic}, Stanford CoreNLP \cite{manning2014stanford}, Apache OpenNLP and a very recent open source collection of Transformer-based models from Hugging Face Inc. \cite{wolf2019transformers}, the spaCy library is optimised for speed on CPUs, has an intuitive API and easily integrates with the active learning-based annotation tool Prodigy \cite{Prodigy:2018}. The architecture of SpaCy's NER model is based on convolutional neural networks with tokens represented as hashed Bloom embeddings \cite{serra2017getting} of prefix, suffix and lemmatisation of individual words augmented with a transition-based chunking model \cite{lample2016neural}. We also experimented with various combinations of hyperparameters of the neural network architecture, dropout rates, batch compounding, learning rate and regularisation schemes. We set aside 30 documents ($10\%$) sampled at random from the training data as a validation set.

\subsubsection{Model training augmentation with bootstrapped noisy labels}

Several recent lines of research have demonstrated a clear benefit in terms of achieving higher accuracy and better generalisation of neural networks trained with corrupted, noisy and synthetically augmented data \cite{xie2019self, natarajan2013learning, provilkov2019bpe, anaby2019not}. Training with data augmentation also alleviates the problem of learning from a limited amount of manually annotated data. Similar to the idea presented in 'Snorkel' \cite{ratner2019snorkel}, we designed a number of labelling functions (LF) by compiling a list of rules and keyword patterns for all seven named-entity categories. Additionally, we exploited a 'sense2vec' approach \cite{trask2015sense2vec} which  was fine-tuned on the entire MIMIC-III corpora to bootstrap keywords and patterns. 'Sense2vec' is a more complex version of the 'Word2vec' method \cite{mikolov2013efficient} for representation of words as vectors. The major improvement over 'word2vec' is that 'sense2vec' also learns from linguistic annotations of words for sense disambiguation in their embeddings. 

The resulting labelling functions were used to created a 'silver' training set consisting of annotated data by string pattern matching. The NER model was then trained by using a combination of gold and silver annotated examples in each batch. In order to prevent data leakage and a biased inflation of the performance metrics, such as precision and recall, the model was tested only on gold annotated data set comprising 202 documents (cf. Table \ref{tab:annotated_summary_stats}) provided by the n2c2 2018 challenge. 

\subsubsection{Model evaluation}
\label{model_evaluation}

In order to estimate the performance of the proposed named-entity recognition model, we used the evaluation schema proposed in SemEval'13 and outlined in \ref{a:evaluation_schema}. The evaluation schema comprised a number of potential errors categories produced by the model and the model performance metrics, such as precision and recall were computed using the expressions \ref{eq:a_1}. Under the current evaluation schema, partial match was considered as an exact match between the gold-annotated and the predicted labels while no restriction was imposed on the boundaries of the tokens. The rationale behind this approach was obvious from the ambiguity in gold-annotations examples corresponding to the same concept. For example, both sequences 'for 3 weeks' and '3 weeks' were labelled as 'Duration'. In particular, 492 of 967 ($71\%$) text spans labelled as 'Duration' started with the word 'for'. 

We estimated both, strict and lenient metrics. Strict metrics accounts only for the exact match in both, surface strings and the corresponding labels, whereas the lenient metrics allow for partial matches. Specifically, strict and lenient metrics were obtained from \ref{eq:a_1} with $\alpha=0$ and $\alpha=1$ correspondingly. We reported both, micro and macro averaged precision and recall and their corresponding F1 scores.



\section{Results}
\label{results}

\subsection{Model pre-training}

\begin{figure}
\begin{floatrow}
\resizebox{\columnwidth}{!}{%
\capbtabbox{%
    \begin{tabular}{l|r|r|r|r}
     Configuration  &   Width   &   Depth   &   Time& Size\\\hline\hline
      (default)     &   96      &   4       &   73  &   3.8\\
                    &   128     &   8       &   90  &   18.3\\
                    &   256     &   8       &   118 &   47.6\\
                    &   256     &   16      &   164 &   66.1\\\hline\hline
     \multicolumn{2}{l}{Number of documents} & \multicolumn{3}{r}{2,083,054}\\
     \multicolumn{2}{l}{Number of words} & \multicolumn{3}{r}{3,129,334,419}\\\hline
    \end{tabular}}
    {\caption{Model pre-training characteristics for various combinations of convolutional layers dimensions. Time and the resulting model size are reported in minutes and megabytes (MB) respectively.}\label{tab:ssl}}
\ffigbox{%
  \includegraphics[width=0.42\textwidth]{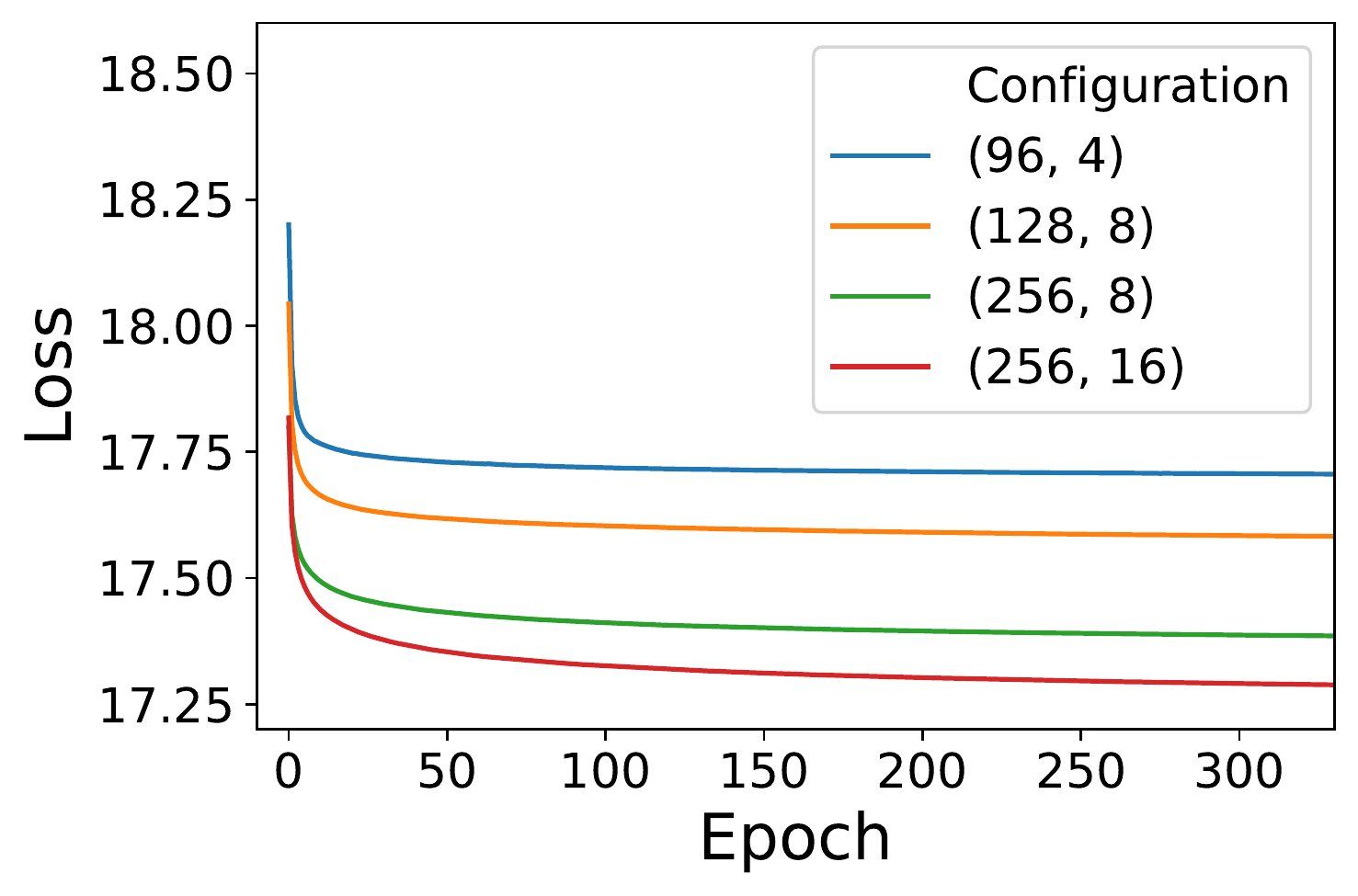}}
  {\caption{The decaying loss of pre-trained models.}\label{fig:ssl}}}
\end{floatrow}
\end{figure}

The pre-training task was performed on the entire MIMIC-III data set for 350 epochs using a number of configurations of the width and depth of the convolutional layers. Each configuration was trained on a single GTX 2080 Ti GPU. CNN dimensions, summary statistics of the pre-training text corpus, the average running time per epoch in minutes and the model size in MB are summarised in Table \ref{tab:ssl}. The corresponding training losses, logarithmically scaled, are plotted in Fig. \ref{fig:ssl}.

\subsection{Rationale for collecting more training data}
\label{rationale}

\begin{wraptable}[]{r}[0pt]{0.45\linewidth}
\centering
\resizebox{\columnwidth}{!}{
\begin{tabular}{r|r|r}
            Fraction   &   Accuracy    &   Delta \\\hline\hline
            0$\%$      &   0.0         &   baseline \\ 
           25$\%$      &   90.66       &   +90.66 \\ 
           50$\%$      &   91.93       &   +1.27 \\ 
           75$\%$      &   92.42       &   +0.49 \\            
          100$\%$      &   92.63       &   +0.21 \\\hline
    \end{tabular}}
    \caption{Change in accuracy with more training data. Delta denotes a relative improvement.}
\label{tab:train-curve}
\end{wraptable}
Generally, collecting more training data will improve the model accuracy and lead to better generalisation. We simulated, using the Prodigy library and 'train-curve' recipe, an acquisition of more data by training of NER model on fractions (25$\%$, 50$\%$, 75$\%$ and 100$\%$) of the training set and evaluating on the test set. We indeed observed (Table \ref{tab:train-curve}) a steady upward trend in improvement of accuracy while using more training data, especially in the last fraction of data which indicates the advantages of further collecting more data.

\subsection{Named-entity recognition model}

The developed Med7 clinical named-entity recognition model was trained in total on 1212 documents, comprising 303 silver training examples augmented with gold annotated data from the official 303 documents from the n2c2 training data (cf. Table \ref{tab:annotated_summary_stats}) and additionally manually gold annotated 606 documents, randomly sampled from discharge letters of MIMIC-III ensuring that there are no documents present from the testing data. The manual annotation was performed using Prodigy, an active learning annotation tool, following the general procedure outlined in \cite{vaci2020natural}. The baseline NER model for the active-learning support containing all seven categories was trained on the official 303 documents. The baseline NER model was used within the Prodigy 'human-in-the-loop' framework to suggest entities on unseen texts and a human annotator accepted or corrected model predictions, creating gold annotated examples. We obtained the inter-annotator agreement F1 score of 0.924 between the gold n2c2 annotations and of our two annotators and F1 score of 0.989 between our annotators. The explicit toke-level confusion matrices along with summary statistics are presented in Table \ref{b:confusion_matrix_gold_annotator_1}, Table \ref{b:confusion_matrix_gold_annotator_2} and Table \ref{b:iaa_scores} accordingly. For generating silver training data, we used spaCy python library for keyword phrase matching with 'EntityRuler' class along with linguistic pattern matching with exemplars from the training data set. Drug names, both generic and brand names, were sourced from publicly available online resources. Training results and evaluation statistics are summarised in Table \ref{tab:training_data} and Table \ref{tab:test_results} correspondingly. A more detailed token-level confusion matrix showing the exact model predictions is summarised in Table \ref{tab:confusion_matrix}.

\begin{table}[!htb]
    \centering
    \begin{tabular}{lrrrr}
                    &   Gold (n2c2)    &   Silver      &   Prodigy     &   Total  \\\hline\hline
        Dosage      &   4227    &   2792        &   3437        &   10456   \\
        Drug        &   16257   &   10551       &   12687       &   39495   \\
        Duration    &   592     &   462         &   620         &   1674    \\
        Form        &   6657    &   4299        &   5056        &   16012   \\
        Frequency   &   6281    &   4317        &   5106        &   15704   \\
        Route       &   5460    &   3761        &   4554        &   13775   \\
        Strength    &   6694    &   4328        &   5246        &   16268   \\\hline\hline
        Number of documents &   303 &   303     &   606         &   1212    \\\hline
    \end{tabular}
    \caption{The distribution of annotated text spans in three data sets used for training of the NER model.}
    \label{tab:training_data}
\end{table}

\begin{table}[!htb]
    \centering
    \begin{tabular}{lccc|ccc}
        \multicolumn{1}{}{} & \multicolumn{3}{c}{Strict}& \multicolumn{3}{c}{Lenient}\\
                    &   Precision   &   Recall  &   F1  &   Precision   &   Recall  &   F1\\\hline\hline
        Dosage      &   0.879   &   0.831   &   0.854   &   0.957   &   0.904   &   0.931\\
        Drug        &   0.954   &   0.926   &   0.941   &   0.984   &   0.956   &   0.971\\
        Duration    &   0.817   &   0.733   &   0.773   &   0.953   &   0.854   &   0.901\\
        Form        &   0.921   &   0.886   &   0.903   &   0.983   &   0.947   &   0.965\\
        Frequency   &   0.801   &   0.784   &   0.792   &   0.989   &   0.969   &   0.979\\
        Route       &   0.961   &   0.943   &   0.952   &   0.973   &   0.954   &   0.964\\
        Strength    &   0.927   &   0.781   &   0.848   &   0.992   &   0.836   &   0.907\\\hline
        Average (micro)       &   0.916   &   0.871   &   0.893 &   0.982   &   0.933   &   0.957\\
        Average (macro)       &   0.897   &   0.844   &   0.869 &   0.977   &   0.919   &   0.947\\\hline
    \end{tabular}
    \caption{The evaluation results of the NER model on the test set with 202 documents.}
    \label{tab:test_results}
\end{table}

\subsection{Translation to UK-CRIS data}

One of the challenges in developing a robust clinical information extraction system, is in its generalisability beyond the data distribution it was trained on. Accurate algorithms developed using data from a small number of medical centres, have demonstrated their poor generalisability when applied within a similar context to other medical centres. For example, in a recent study on the algorithmic approach to early detection of sepsis \cite{reyna2020early}, the training data were sourced from electronic health records of two hospitals, while the data from a third hospital were used for testing the developed algorithm. It has been demonstrated and discussed in details \cite{morrill2019signature} that a highly accurate predictive algorithm, validated on a fraction of data from the same two hospitals, failed to achieve the same level of accuracy when tested on the data from the third hospital, not included in the training process. Poor performance using the out-of-distribution (OOD) data poses a significant challenge on wider applications of the developed models and is highly important when algorithms inform real-world decisions \cite{ren2019likelihood}. 

    \begin{wraptable}[]{r}[0pt]{0.5\linewidth}
    \resizebox{\columnwidth}{!}{%
    \begin{tabular}{l|r|r|r}
        Clinical concepts   &   Train   &   Test    &   Total   \\ \hline\hline
        Dosage          &   298    &   48    &   346    \\
        Drug            &   3253   &   571   &   3824   \\
        Duration        &   1006     &   215     &   1221     \\        
        Form            &   410    &   63    &   473   \\
        Frequency       &   1604    &   305    &   1909   \\
        Route           &   208    &   32    &   240    \\
        Strength        &   1338    &   276    &   1614   \\\hline        
        Number of texts &   536   &   134     &   670     \\\hline
    \end{tabular}}
    \caption{Distribution of gold-annotated entities and text summary statistics of the OxCRIS training and test data sets. The number of unique tokens is computed by lowercasing words.}
    \label{tab:annotated_summary_stats_oxcris}
    \end{wraptable}
We investigated how accurate the developed Med7 model, trained on MIMIC-III electronic health records sourced from the Beth Israel Deaconess Medical Center in Massachusetts (United States), can be when applied to CRIS electronic health records in the United Kingdom. We selected a random sample of 670 documents from the Oxford Health NHS Foundation Trust (OFHT) instance of UK-CRIS Network and asked a clinician to annotate them for seven categories following the official guidelines of the n2c2 challenge. 
 
The token-level confusion matrix and the performance metrics of the Med7 model trained on n2c2 data from MIMIC-III and applied to CRIS data from Oxford instance are presented in Table \ref{tab:cm_med7_ox_cris_before} and in Table \ref{tab:med7_ox_cris_metrics} correspondingly. Direct comparison to the results presented in Table \ref{tab:test_results} (F1=0.762 vs. F1=0.944) clearly shows the problem of direct transferability of NER models trained on different data sources.

\begin{table}[!htb]
    \centering
    \resizebox{\columnwidth}{!}{%
    \begin{tabular}{lccc|ccc}
        \multicolumn{1}{}{} & \multicolumn{3}{c}{Before fine-tuning on OxCRIS}& \multicolumn{3}{c}{After fine-tuning on OxCRIS}\\\cline{2-7}
                    &   Precision   &   Recall  &   F1  &   Precision   &   Recall  &   F1\\\hline\hline
        Dosage      &   0.826   &   0.396   &   0.535   &   0.656   &   0.833   &   0.734\\
        Drug        &   0.912   &   0.968   &   0.939   &   0.975   &   0.977   &   0.976\\
        Duration    &   0.951   &   0.107   &   0.192   &   0.883   &   0.934   &   0.908\\
        Form        &   0.554   &   0.611   &   0.581   &   0.924   &   0.968   &   0.946\\
        Frequency   &   0.912   &   0.332   &   0.487   &   0.941   &   0.944   &   0.942\\
        Route       &   0.348   &   0.719   &   0.469   &   0.882   &   0.938   &   0.909\\
        Strength    &   0.938   &   0.877   &   0.906   &   0.996   &   0.917   &   0.955\\\hline
        Average (micro)       &   0.864   &   0.681   &   0.762 &   0.941   &   0.947   &   0.944\\
        Average (macro)       &   0.778   &   0.586   &   0.609 &   0.901   &   0.932   &   0.914\\\hline
    \end{tabular}}
    \caption{The lenient evaluation results of the Med7 model using 134 test documents sourced from OxCRIS - the Oxford Health NHS Foundation Trust from within the UK-CRIS electronic health records Network.}
    \label{tab:med7_ox_cris_metrics}
\end{table}


\section{Discussion}
\label{discussion}

The developed named-entity recognition model for clinical concepts in unstructured medical records was trained to recognise seven categories, such as drug names, including both generic and brand names, dosage of the drugs, their strength, the route of administration, prescription duration and the frequency. The data for model development and testing was sourced from the n2c2 challenge, comprising a collection of 303 and 202 documents for training and testing respectively,  which represent a sample from the MIMIC-III electronic health records. We demonstrated (Section \ref{rationale}) that collecting more annotated examples would improve the model accuracy and therefore implemented two approaches for obtaining more annotations: noisy labelling and active learning with 'human-in-the-loop'. For the noisy labelling, we create a list of unique patterns for each of the seven categories, sourced from the training corpus and from external resources available on the internet, and then used regular expression with string pattern matching to assign labels to tokens. Our two annotators were trained by closely following the official 2018 n2c2 annotation guidelines and demonstrated a high level of inter-annotator agreement among themselves (F1=0.989) as well as a high-level of concordance (F1=0.924) with the gold-annotations provided by the organisers of 2018 n2c2 Challenge (cf. Table \ref{b:iaa_scores}).

The overall (micro-averaged) performance of the NER model across all seven categories was F1=0.957 (0.893), with Precision=0.982 (0.916) and Recall=0.933 (0.871) for lenient (strict) estimates. More detailed breakdown of the performance for each of the categories is presented in Table \ref{tab:test_results}. The performance for 'Duration' and 'Frequency' categories was poorer. There were intrinsically fewer cases of 'Duration' ($\sim$ 1.5$\%$) appeared in texts and these concepts were also ambiguously annotated as mentioned in Section  \ref{model_evaluation}. A similar situation was also observed for the 'Frequency' category, where in spite of a good number of the annotated examples ($\sim$ 14$\%$), the ambiguity in the presentation of text spans was higher, which resulted in a large number of partial matches (cf. Table \ref{tab:confusion_matrix}). Another reason for poor performance for both 'Duration' and 'Frequency' was due to inconsistent annotations, where the same text string appeared in both categories.

Self-supervised pre-training of deep learning models has shown its efficiency in many NLP task. We experimented with a number of architectural variations of the width and depth of convolutional layers as well as the size of the embedding rows. Empirically, and as confirmed by other studies \cite{raffel2019exploring}, larger models, with more parameters, tend to achieve better results. Interestingly, the larger model (Width=256, Depth=16, Embeddings=10000) outperformed the default one (Width=96, Depth=4, Embeddings=2000) by a small margin (F1$_{256}$=0.893 vs F1$_{96}$=0.884) however, the differences were more visible for 'Duration' (F1$_{256}$=0.773 vs F1$_{96}$=0.729) and 'Strength' (F1$_{256}$=0.848 vs F1$_{96}$=0.801). The better performance resulted at the expense of the training time, its size on a disk and the memory consumption. We publicly released the pre-trained neural network weights for various architectures through the dedicated GitHub repository\footnote{https://github.com/kormilitzin/med7}.

Another objective of this work was to estimate the degree of transferability of the developed information extraction model to another clinical domain. We evaluated how the Med7 model, trained on a collection of discharge letters from the intensive care unit in the US (MIMIC), performed on the secondary care mental health medical records in the UK (CRIS). The Med7 model was purposely designed to recognise non-context related medical concepts, such as drug names, strength, dosage, duration, route, form and frequency of administration and we expected to see a comparable level of the model performance across the both EHR systems. To consistently validate the transferability of the Med7 model, a random sample of 670 gold-annotated examples from OxCRIS were split into training (536) and test (134) data sets (cf. Table \ref{tab:annotated_summary_stats_oxcris}). We compared the performance of the Med7 model without and with fine-tuning on OxCRIS. The direct application of Med7 on the testing set of 134 documents, resulted in a quite poor performance (F1=0.762). We investigated the cases where the model was predicting incorrectly and in the majority of them, the main reason for poor performance was due to differences in the language presentation of the concepts. For examples, the model largely missed concepts labelled as 'Frequency' in OxCRIS, such as "ON", ("every night"), "OD" ("every day"), "BD" ("twice daily"), "OM" ("every morning"), "mane" and "nocte". Then, we fine-tuned the Med7 model on the training set of OxCRIS (536 documents) and evaluated on the same testing set as before of 134 documents. Despite the small number of training examples in OxCRIS, leveraging the transfer learning approach of re-using the pre-trained Med7 model on MIMIC, resulted in higher accuracy (F1=0.944) comparable with training and testing on the same domain (cf. Table \ref{tab:med7_ox_cris_metrics}).

One strength on this project is in the interoperability of the developed model with other generic deep learning NLP libraries, such as HuggingFace and Thinc as well as straightforward integration with pipelines developed under the spaCy framework. This allows to customise the Med7 model and include other pipeline components, such as negation detection, entity relations extraction and to map the extracted concepts onto the universal medical language system (UMLS). Normalisation of concepts to UMLS categories will allow to systematically parse electronic medical records into structured and consistent tabular form which will be ready for downstream epidemiological analyses. Additionally, the developed model naturally integrates into the Prodigy annotation tool, which allows to efficiently collect more gold-annotated examples. It is also worth mentioning that the Med7 model is designed to run on standard CPUs, rather than expensive GPUs. This fact will allow researchers without access to expensive and complex infrastructure to develop fast and robust pipelines for clinical natural language processing.

However, two limitations should be noted. First, is that some of the categories are naturally underrepresented which impacts the accuracy of the NER model. It was observed empirically that the number of annotated 'Duration' entities was intrinsically skewed in the medical records, in contrast to drug names and strength, making it more challenging to train a robust model to accurately identify these entities. Interestingly, the same pattern of the number of reported mentions of the 'Duration' category persists in both, MIMIC and OxCRIS data, which might be indicative of a general clinical reporting pattern. A second limitation of this study is related to a low number of the manually-annotated examples in OxCRIS, in order to run more rigours evaluations of the transferability of the Med7 model across all seven categories. 

Future research into the robust clinical information extraction system will need to further address the feasibility of deploying the model in the UK-CRIS Network Trust members and evaluate its transferability. The aim is to furnish clinical researchers with an open source and a robust tool for structuring free-text patients' data for downstream analytical tasks.

\section{Conclusion}
\label{conclusion}

In this work we developed and validated a clinical named-entity recognition model for free-text electronic health records. The model was developed using the MIMIC-III free-text data and trained on a combination of the manually annotated data from the 2018 n2c2 challenge, on a random sample from MIMIC-III with noisy labels and manually annotated data using active learning with Prodigy. To maximise the utilisation of a large amount of unstructured free-text data and alleviate the problem of training from limited data, we used self-supervised learning to pre-train the weights of the NER neural network model. We demonstrated that transfer learning plays an essential role in developing a robust model applicable across different clinical domains and the developed Med7 model does not require an expensive infrastructure and can be used on standard machines with CPU. Further research is needed to improve recognition of naturally underrepresented concepts and we are planning to address this problem, as well as extracted concepts normalisation and UMLS linkage in our future releases of the Med7 model.

\section*{Acknowledgments}

The study was funded by the National Institute for Health Research's (NIHR) Oxford Health Biomedical Research Centre (BRC-1215-20005). This work was supported by the UK Clinical Records Interactive Search (UK-CRIS) system funded and developed by the NIHR Oxford Health BRC at Oxford Health NHS Foundation Trust and the Department of Psychiatry, University of Oxford. AK, NV, QL, ANH are funded by the MRC Pathfinder Grant (MC-PC-17215). We are thankful to the organisers of the n2c2 2018 Challenge for providing annotated corpus and the annotation guidelines.

The views expressed are those of the authors and not necessarily those of the UK National Health Service, the NIHR, or the UK Department of Health.

We would also like to acknowledge the work and support of the Oxford CRIS Team: Tanya Smith, Head of Research Informatics and Biomedical Research Centre (BRC) CRIS Theme Lead and Lulu Kane, Adam Pill and Suzanne Fisher, CRIS Academic Support and Information Analysts.

\appendix

\section{The evaluation schema for extracted concepts}
\label{a:evaluation_schema}
In order to evaluate the output of the NER system, we adopted the notations developed for different categories of errors \cite{chinchor-sundheim-1993-muc} and the evaluation schema introduced in SemEval'13 (cf. Eq.\ref{eq:a_1}). The following types of evaluation errors were considered (Table \ref{tab:eval_schema}):

\begin{table}[!htb]
    \centering
    \resizebox{\columnwidth}{!}{
    \begin{tabular}{l|ll|cc|cc}
           \multicolumn{3}{c}{Error Type}  &   \multicolumn{2}{c}{Gold Standard}   &   \multicolumn{2}{c}{NER Prediction}  \\\hline\hline
        &   \multicolumn{2}{c}{} &  \multicolumn{1}{|c}{Text span}   &   \multicolumn{1}{c}{Label}   &   \multicolumn{1}{|c}{Text span}   &   \multicolumn{1}{c}{Label}           \\\cline{4-7}
        1   &  Correct     &   (COR)   &   aspirin     &   Drug    &   aspirin     &   Drug            \\
        2   &   Incorrect   &   (INC)   &   25        &   Strength    &   25     &   Dosage            \\
        3   &   Partial     &   (PAR)   &  Augmentin        &   Drug        & Augmentin XR &   Drug    \\
        4   &   Partial     &   (PAR)   &  for 3 weeks        &   Duration        & 3 weeks &   Duration    \\
        5   &   Partial     &   (PAR)   &  p.r.n.        &   Frequency        & prn &   Frequency    \\
        6   &   Missing     &   (MIS)   &  tablet    &   Form    &    \multicolumn{1}{c}{-}       &      \multicolumn{1}{c}{-}         \\
        7   &   Spurious    &   (SPU)   &      \multicolumn{1}{c}{-}        &     \multicolumn{1}{c|}{-}      & Codeine   & Drug                  \\\hline
    \end{tabular}}
    \caption{A list of examples of typical errors produced by the NER model.}
    \label{tab:eval_schema}
\end{table}

where Correct(COR) represents a complete match of both, the annotation boundary and the entity type. Incorrect(INC) is the case where at least one of the predicted boundary or the entity type do not match. Partial(PAR) match corresponds to predicted entity boundary which overlaps with ground-truth annotation, but they are not exactly the same. Missing(MIS) the case where the ground-truth annotated boundary is not predicted by the NER, but the ground-truth string is present in the gold-annotated corpus. Spurious(SPU) corresponds to predicted entity boundary which does not exist in the gold-annotated corpus.

\begin{equation} \label{eq:a_1}
    \begin{split}
    \text{Possible (POS)}   &= COR+INC+PAR+MIS=TP+FN \\
    \text{Actual (ACT)}     &= COR+INC+PAR+SPU=TP+FP \\
    \text{Precision}        &=   (COR + \alpha PAR) / ACT\\
    \text{Recall}           &=   (COR + \alpha PAR) / POS
    \end{split}
\end{equation}


The detailed token-level confusion matrix of the model predictions following the error types defined in \ref{eq:a_1}.

\begin{table}[!htb]
    \centering
    \resizebox{\columnwidth}{!}{
    \begin{tabular}{cr||rrrrrrr|rr}
         \multicolumn{2}{c}{} & \multicolumn{7}{c}{\textbf{Predicted categories}} & \multicolumn{1}{c}{}\\
                &    &   Dosage  &   Drug    &   Duration    &   Form    &   Frequency   &   Route   &   Strength    &   Missed  & Partial \\\hhline{~==========}
    \parbox[t]{3mm}{\multirow{7}{*}{\rotatebox[origin=c]{90}{\textbf{True categories}}}}     
         & Dosage     &   2225    &   0         &       6       &       10  &       24       &       1   &       16      &   200     & 199   \\
         & Drug       &   2       &   9796      &       0       &       7  &       0       &       4   &        1      &   449     & 316   \\
         & Duration   &   6       &   0         &       277     &       0   &       8       &       0   &        2       &   39     & 46    \\
         & Form       &   38      &   31        &       0       &    3864   &       1       &       65  &        6       &   90     & 264 \\
         & Frequency  &   1       &   3         &       4       &       5   &       3144    &       2   &        0       &   108     & 745  \\
         & Route      &   3       &   4         &       0       &       43  &       1       &       3312&        1        &  108     & 41  \\
         & Strength   &   38      &   3         &       0       &       1  &       2       &       0   &       3304    &    650     & 232  \\\cline{2-11}
         & Spurious   &   20      &   120       &       6      &        4    &       7       &       22  &       3       &          \\
    \end{tabular}}
    \caption{Token-level confusion matrix of the predicted entities versus the ground truth labels. Spurious examples correspond to predicted entity boundary and type which do not exist in ground-truth annotations and partial matches correspond to predicted entity boundary overlap with golden annotation, but they are not the same. Missing entities correspond to ground-truth annotation boundary which were not identified.}
    \label{tab:confusion_matrix}
\end{table}

\section{Inter-annotator agreement analysis}
\label{b:iaa}

We estimated the level of concordance between the gold-annotated corpus from the n2c2 2018 challenge and two trained annotators. The annotators closely followed the same annotation guidelines as used in the challenge. Ten documents were sampled at random from 202 documents comprising the test set. The distribution of gold-annotated tokens and by two annotators is presented in Table \ref{b:iaa_stats}.

\begin{table}[!htb]
    \centering
    \begin{tabular}{l|r|r|r}
        Types of annotated entities   &   Gold (n2c2)   &   Annotator 1    &   Annotator 2   \\ \hline\hline
        Dosage          &   128    &   139    &   139    \\
        Drug            &   519   &   530   &   526   \\
        Duration        &   28     &   31     &   32     \\        
        Form            &   234    &   246    &   238   \\
        Frequency       &   193    &   196    &   201   \\
        Route           &   179    &   167    &   167    \\
        Strength        &   200    &   212    &   205   \\\hline        
        Number of documents & 10   &   10     &   10     \\\hline
    \end{tabular}
    \caption{The number of the gold and manually annotated entities for the inter-annotator agreement evaluation corpus, comprising ten randomly sampled texts from the test set of 202 documents.}
    \label{b:iaa_stats}
\end{table}

\begin{table}[!!htb]
    \centering
    \resizebox{\columnwidth}{!}{
    \begin{tabular}{cr||rrrrrrr|rr}
         \multicolumn{2}{c}{} & \multicolumn{7}{c}{\textbf{Annotator 1}} & \multicolumn{1}{c}{}\\
                &    &   Dosage  &   Drug    &   Duration    &   Form    &   Frequency   &   Route   &   Strength    &   Missed  & Partial \\\hhline{~==========}
    \parbox[t]{3mm}{\multirow{7}{*}{\rotatebox[origin=c]{90}{\textbf{Gold (n2c2)}}}}     
         & Dosage     &   104    &   0       &       1       &       3   &       0       &       0   &       2      &   17   & 4   \\
         & Drug       &   0       &   473   &       0       &       3   &       0       &       1   &       0      &   27     & 21   \\
         & Duration   &   0       &   0       &       19     &       0   &       0       &       0   &       0       &   2    & 7    \\
         & Form       &   1       &   4       &       0       &   201    &       0       &       2  &       0       &   7   & 21 \\
         & Frequency  &   1       &   0       &       0       &       0   &       172    &       0   &       1       &   2   & 17  \\
         & Route      &   2       &   2       &       0       &   2      &       0       &       156&       0       &   15   & 2  \\
         & Strength   &   2      &   1       &       0       &       0   &       0       &       0   &       171    &   4   & 28  \\\cline{2-11}
         & Spurious   &   25     &   29     &       4      &       16 &        7     &       6 &       10      &          \\
    \end{tabular}}
    \caption{Token-level confusion matrix of the annotator 1 versus the gold-standard annotations provided by 2018 n2c2 challenge.}
    \label{b:confusion_matrix_gold_annotator_1}
\end{table}

\begin{table}[!htb]
    \centering
    \resizebox{\columnwidth}{!}{
    \begin{tabular}{cr||rrrrrrr|rr}
         \multicolumn{2}{c}{} & \multicolumn{7}{c}{\textbf{Annotator 2}} & \multicolumn{1}{c}{}\\
                &    &   Dosage  &   Drug    &   Duration    &   Form    &   Frequency   &   Route   &   Strength    &   Missed  & Partial \\\hhline{~==========}
    \parbox[t]{3mm}{\multirow{7}{*}{\rotatebox[origin=c]{90}{\textbf{Gold (n2c2)}}}}     
         & Dosage     &   104    &   0       &       1       &       3   &       0       &       0   &       2      &   17   & 4   \\
         & Drug       &   0       &   472   &       0       &       3   &       0       &       1   &       0      &   30     & 20   \\
         & Duration   &   0       &   0       &       19     &       0   &       0       &       0   &       0       &   2    & 7    \\
         & Form       &   0       &   3       &       0       &   201    &       0       &       2  &       0       &   9   & 21 \\
         & Frequency  &   0       &   0       &       1       &       0   &       172    &       0   &       0       &   2   & 18  \\
         & Route      &   2       &   2       &       0       &   2      &       0       &       156 &       0       &   15   & 2  \\
         & Strength   &   3      &   1       &       0       &       0   &       4       &       0   &       171    &   3   & 21  \\\cline{2-11}
         & Spurious   &   26     &   28     &       4      &       8 &        7     &       6 &       10      &          \\
    \end{tabular}}
    \caption{Token-level confusion matrix of the annotator 2 versus the gold-standard annotations provided by 2018 n2c2 challenge}
    \label{b:confusion_matrix_gold_annotator_2}
\end{table}

\begin{table}[!htb]
    \centering
    \resizebox{\columnwidth}{!}{%
    \begin{tabular}{lccc|ccc|ccc}
        \multicolumn{1}{}{} & \multicolumn{3}{c}{Annot. 1 vs. Gold} & \multicolumn{3}{|c}{Annot. 2 vs. Gold}& \multicolumn{3}{|c}{Annot. 1 vs. Annot. 2}\\\cline{2-10}
                    &   Pr      &   Re      &   F1      &   Pr      &   Re      &   F1      &   Pr      &   Re      &   F1\\\hline\hline
        Dosage      &   0.777   &   0.824   &   0.801   &   0.777   &   0.824   &   0.801   &   0.986   &   0.986   &   0.986\\
        Drug        &   0.935   &   0.935   &   0.935   &   0.935   &   0.935   &   0.935   &   0.998   &   0.991   &   0.994\\
        Duration    &   0.812   &   0.923   &   0.867   &   0.812   &   0.929   &   0.867   &   0.969   &   1.000   &   0.984\\
        Form        &   0.933   &   0.941   &   0.937   &   0.933   &   0.941   &   0.937   &   1.000   &   0.967   &   0.983\\
        Frequency   &   0.945   &   0.984   &   0.964   &   0.945   &   0.984   &   0.964   &   0.975   &   1.000   &   0.987\\
        Route       &   0.946   &   0.883   &   0.913   &   0.946   &   0.883   &   0.913   &   1.000   &   1.000   &   1.000\\
        Strength    &   0.941   &   0.946   &   0.944   &   0.941   &   0.946   &   0.944   &   1.000   &   0.962   &   0.981\\\hline
        Average (micro)       &   0.921   &   0.928   &   0.924 &   0.921   &   0.928   &   0.924   &   0.994   &   0.985   &   0.989\\
        Average (macro)       &   0.901   &   0.921   &   0.911 &   0.901   &   0.921   &   0.911   &   0.991   &   0.986   &   0.988\\\hline
    \end{tabular}}
    \caption{The evaluation results of the inter-annotator agreement on a random selection of ten documents from the 202 test texts. A pair-wise comparison between each of the annotators and the gold-annotated documents as well as the direct comparison between the both annotators.}
    \label{b:iaa_scores}
\end{table}

We examined the cases where our two annotators labelled the concepts of interests differently than those found in the gold-annotated data set provided by the n2c2 team.


\section{Fine-tuning on UK-CRIS}

\begin{table}[!htb]
    \centering
    \resizebox{\columnwidth}{!}{
    \begin{tabular}{cr||rrrrrrr|rr}
         \multicolumn{2}{c}{} & \multicolumn{9}{c}{\textbf{Med7-predicted categories: \underline{before} fine-tuning on OxCRIS}} \\
                &    &   Dosage  &   Drug    &   Duration    &   Form    &   Frequency   &   Route   &   Strength    &   Missed  & Partial \\\hhline{~==========}
    \parbox[t]{3mm}{\multirow{7}{*}{\rotatebox[origin=c]{90}{\textbf{Gold annotated}}}}     
         & Dosage     &   18        &   0       &       0       &       0       &       0       &       0   &       12      &   17      & 1   \\
         & Drug       &   0         &   535     &       0       &       0       &       0       &       0   &       0       &   18      & 15   \\
         & Duration   &   0         &   0       &       18      &       0       &       1       &       0   &       0       &   158     & 1    \\
         & Form       &   0         &   2       &       0       &       34      &       0       &       1   &       0       &   20      & 2 \\
         & Frequency  &   0         &   7       &       0       &       25      &       86      &       40  &       1       &   114     & 7  \\
         & Route      &   0         &   0       &       0       &       3       &       3       &       23  &       0       &   6       & 0  \\
         & Strength   &   3         &   0       &       0       &       0       &       0       &       0   &       238     &   31      & 4  \\\cline{2-11}
         & Spurious   &   1         &   44      &       1       &       1       &       8       &       2   &       3       &          \\
    \end{tabular}}
    \caption{Token-level confusion matrix of the Med7 model trained on MIMIC-III and applied to 134 manually annotated documents from the Oxford instance (OxCRIS) of the UK-CRIS electronic medical records Network.}
    \label{tab:cm_med7_ox_cris_before}
\end{table}

\begin{table}[!htb]
    \centering
    \resizebox{\columnwidth}{!}{
    \begin{tabular}{cr||rrrrrrr|rr}
         \multicolumn{2}{c}{} & \multicolumn{9}{c}{\textbf{Med7-predicted categories: \underline{after} fine-tuning on OxCRIS}} \\
                &    &   Dosage  &   Drug    &   Duration    &   Form    &   Frequency   &   Route   &   Strength    &   Missed  & Partial \\\hhline{~==========}
    \parbox[t]{3mm}{\multirow{7}{*}{\rotatebox[origin=c]{90}{\textbf{Gold annotated}}}}     
         & Dosage     &   39        &   0       &       0       &       0       &       0       &       0   &       1       &   7       & 1     \\
         & Drug       &   0         &   553     &       0       &       2       &       0       &       0   &       0       &   11      & 4     \\
         & Duration   &   0         &   0       &       177     &       0       &       1       &       0   &       0       &   13      & 20    \\
         & Form       &   0         &   0       &       0       &       61      &       1       &       1   &       0       &   0       & 0     \\
         & Frequency  &   1         &   1       &       0       &       2       &       279     &       1   &       0       &   12      & 6     \\
         & Route      &   0         &   0       &       0       &       0       &       0       &       30  &       0       &   2       & 0     \\
         & Strength   &   16        &   1       &       0       &       0       &       0       &       0   &       242     &   6       & 11    \\\cline{2-11}
         & Spurious   &   4         &   12      &       26      &       1       &       16      &       2   &       0       &                   \\
    \end{tabular}}
    \caption{Token-level confusion matrix of the Med7 model trained on MIMIC-III and applied to 134 manually annotated documents from the Oxford instance (OxCRIS) of the UK-CRIS electronic medical records Network.}
    \label{tab:cm_med7_ox_cris_after}
\end{table}



\newpage
\bibliographystyle{elsarticle-num}
\bibliography{refs.bib}





\end{document}